\documentclass[12pt]{article}

\usepackage{amsmath,amssymb,amscd,amsthm}
\usepackage{latexsym}
\usepackage{bbm}

\usepackage{graphicx}
\usepackage{float}
\usepackage{multirow}
\usepackage[font=small,labelfont=bf]{caption}
\usepackage{subcaption}
\usepackage{adjustbox}
\usepackage{color,xcolor}
\usepackage{array}

\usepackage{algorithm}
\usepackage{algpseudocode}

\usepackage{cite}
\usepackage{hyperref}

\usepackage[left=2cm,top=2.5cm,right=2.5cm,bottom=1.5cm]{geometry}
\usepackage{microtype}

\usepackage{epstopdf}

\usepackage[normalem]{ulem}
\useunder{\uline}{\ul}{}
\usepackage{enumitem}
\setlist[itemize]{align=parleft,left=0pt..1em}

\title{\bfseries Weakly Semi-supervised Whole Slide Image Classification by Two-level Cross Consistency Supervision}

\author{
Linhao Qu$^{1,2,\dagger}$,
Shiman Li$^{1,2,\dagger}$,
Xiaoyuan Luo$^{1,2}$,
Shaolei Liu$^{1,2}$,\\
Qinhao Guo$^{3,4}$,
Manning Wang$^{1,2,\ast}$,
Zhijian Song$^{1,2,\ast}$
}

\date{
\small
$^1$Digital Medical Research Center, School of Basic Medical Science, Fudan University, Shanghai, China\\
$^2$Shanghai Key Lab of Medical Image Computing and Computer Assisted Intervention, Shanghai, China\\
$^3$Department of Gynecologic Oncology, Shanghai Cancer Center, Fudan University, Shanghai, China\\
$^4$Department of Oncology, Shanghai Medical College, Fudan University, Shanghai, China\\
$^\dagger$Equal contribution \quad $^\ast$Corresponding authors: \texttt{zjsong@fudan.edu.cn}
}

\begin{document}

\maketitle

\begin{abstract}
Computer-aided Whole Slide Image (WSI) classification has the potential to enhance the accuracy and efficiency of clinical pathological diagnosis. It is commonly formulated as a Multiple Instance Learning (MIL) problem, where each WSI is treated as a bag and the small patches extracted from the WSI are considered instances within that bag. However, obtaining labels for a large number of bags is a costly and time-consuming process, particularly when utilizing existing WSIs for new classification tasks. This limitation renders most existing WSI classification methods ineffective. To address this issue, we propose a novel WSI classification problem setting, more aligned with clinical practice, termed Weakly Semi-supervised Whole slide image Classification (WSWC). In WSWC, a small number of bags are labeled, while a significant number of bags remain unlabeled. The MIL nature of the WSWC problem, coupled with the absence of patch labels, distinguishes it from typical semi-supervised image classification problems, making existing algorithms for natural images unsuitable for directly solving the WSWC problem. In this paper, we present a concise and efficient framework, named CroCo, to tackle the WSWC problem through two-level Cross Consistency supervision. CroCo comprises two heterogeneous classifier branches capable of performing both instance classification and bag classification. The fundamental idea is to establish cross-consistency supervision at both the bag-level and instance-level between the two branches during training. Extensive experiments conducted on four datasets demonstrate that CroCo achieves superior bag classification and instance classification performance compared to other comparative methods when limited WSIs with bag labels are available. To the best of our knowledge, this paper presents for the first time the WSWC problem and gives a successful resolution.
\end{abstract}

\vspace{1em}

\section{Introduction}
Computer-assisted pathology Whole Slide Image (WSI) classification plays a crucial role in enhancing the accuracy and efficiency of clinical pathology diagnosis by providing essential auxiliary information for clinical decision-making \cite{24,117,118,119,120}. Unlike natural images, WSIs typically have gigapixel sizes, necessitating their division into numerous small patches for processing by deep learning models. However, assigning detailed labels to these patches proves to be highly costly, rendering patch-level supervised learning infeasible \cite{26,27,34,35,37,111}. Multiple Instance Learning (MIL), a common weakly supervised learning paradigm, has emerged as the primary approach for WSI classification. In the weakly supervised MIL setting, each WSI is considered a bag, and the small patches derived from it are regarded as instances within that bag. Positive bags contain at least one positive instance, while negative bags consist entirely of negative instances. Training data only provides bag labels, with no available instance labels \cite{3,4}. In clinical practice, WSI classification serves two primary objectives: bag-level classification, accurately predicting the category of the target bag, and instance-level classification, precisely identifying positive instances within bags predicted as positive. Existing methods for WSI classification can be categorized as bag-based or instance-based, and a brief review of these methods is presented in Section \ref{sec21}.

\begin{figure*}[t!]
  \centering
  \includegraphics[width=\textwidth]{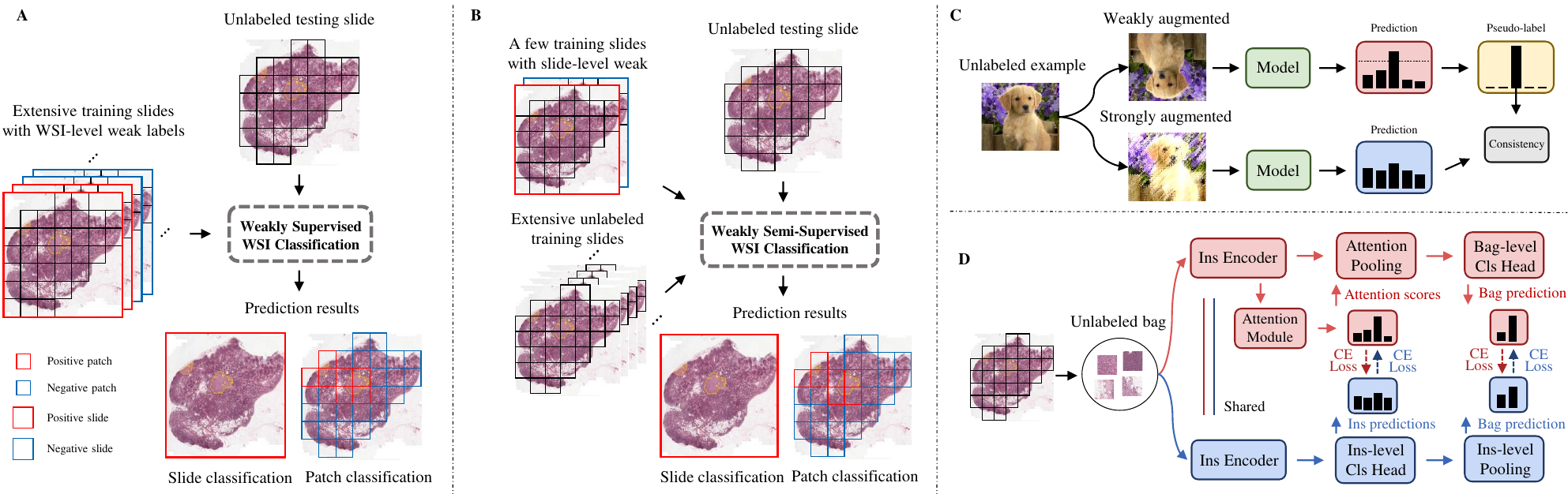}
  \caption{A. The current paradigm for weakly supervised WSI classification assumes the availability of labels for all training bags.
  B. In our proposed weakly semi-supervised WSI classification paradigm, only a limited number of labeled training bags are available, while there are numerous unlabeled bags.
  C. The traditional semi-supervised learning paradigm relies on pseudo-labels and consistency in the domain of natural images.
  D. We introduce CroCo, a concise and efficient framework for bag-level and instance-level Cross Consistency Supervision in WSI classification.}
  \label{figure1}
\end{figure*}

However, providing bag labels for WSIs is a time-consuming and sometimes costly process \cite{24,25,26,27,28} due to the large size of WSIs and the expertise required for labeling. Typically, an experienced pathologist needs to meticulously examine a WSI slide for more than 20-30 minutes. Consequently, it is common to encounter a scenario where there is a substantial number of unlabeled bags and only a limited number of labeled bags for model training, particularly when utilizing existing WSI images for new classification tasks. Although the semi-supervised learning paradigm, which leverages both labeled and unlabeled data for model training, has been extensively studied in natural image classification, the MIL setting of WSI classification sets it apart from typical semi-supervised learning problems. \textbf{In this paper, we present a novel WSI classification problem setting that aligns more closely with clinical practice: Weakly Semi-supervised Whole Slide Image Classification (WSWC).} In WSWC, the training set consists of only a few known positive and negative bag labels, while the labels for the remaining bags are unknown. The objective is to perform both bag classification and instance classification. Figure \ref{figure1} A and B illustrate the existing weakly supervised WSI classification setting and the introduced WSWC setting, respectively.

Semi-supervised learning is a well-established field \cite{55,56,57,81,59,60,67,77,78,79,121,122,123}, as illustrated in Figure \ref{figure1} C, which depicts the classic semi-supervised learning paradigm based on pseudo-labels and consistency. Although it is possible to adapt existing state-of-the-art semi-supervised methods to the WSWC problem by imposing consistency on bag labels, this approach fails to fully leverage the instance-level information that recent studies have shown to be crucial for WSI classification \cite{3,12,18,44}. To address this issue, we design a concise and efficient two-level Cross Consistency supervision framework called CroCo, as shown in Figure \ref{figure1} D. 
CroCo is a typical dual-branch architecture comprising an upper branch, representing the bag-based WSI classifier, and a lower branch, representing the instance-based WSI classifier. Both branches share the same instance encoder.
In the upper branch, an attention-based pooling function is employed to aggregate instance features and derive bag features. Subsequently, a bag-level classification head is trained to make predictions for bags. The attention scores associated with each instance enable instance-level classification by measuring the probability of positivity for each instance. Conversely, the lower branch primarily trains an instance-level classification head by assigning pseudo-labels to each instance. The predicted results are then aggregated to obtain bag-level predictions.

The primary objective of CroCo is to establish a two-level cross-consistency (bag-level and instance-level) between the two heterogeneous branches during training. Specifically, \textbf{for labeled bags}, both branches employ their true labels to conduct bag-level supervision. For instances from labeled positive bags, the attention module in the upper branch generates normalized attention scores, which are then employed as pseudo-labels for the instance classifier in the lower branch. Simultaneously, the instance predictions of the lower branch serve as pseudo-labels for the attention module, thereby achieving cross-supervision between the dual branches at the instance level. Instances from labeled negative bags, known to be negative, are directly employed to train the instance classifier in the lower branch. \textbf{Unlabeled bags} receive cross-consistency supervision at both the bag level and the instance level from both branches. In summary, CroCo offers the following \textbf{key advantages}: \textbf{1) the implementation of a two-level semi-supervised consistency design that maximizes the utilization of instance-level data}, and \textbf{2) the adoption of a heterogeneous dual-branch design that synergistically combines the strengths of diverse models.}

\textbf{The main contributions of this paper are as follows:}

\noindent$\bullet $ We introduce a novel task of Weakly Semi-supervised WSI Classification (WSWC). This task is more aligned with clinical practice than the current WSI classification problem setting that requires supervision of all bag labels.

\noindent$\bullet $ We design a concise and efficient framework to address the WSWC problem by enforcing two-level cross consistency supervision, which is denoted as CroCo. CroCo is based on two heterogeneous classifier branches, and its main idea is to establish both bag-level and instance-level cross-consistency between the two branches. 

\noindent$\bullet $ We thoroughly evaluate CroCo on a synthetic dataset and three real pathological WSI datasets. The results demonstrate that CroCo achieves superior performance in both bag and instance classification compared to strong baselines.

\section{Related Work}
\subsection{Multiple Instance Learning for WSI Classification.}
\label{sec21}
Current MIL methods can be primarily categorized into two types: instance-based methods \cite{1,2,3,28,29,50,109} and bag-based methods \cite{8,9,11,12,13,14,15,30,37,42,45,51,52,110}. Instance-based methods primarily train an instance classifier using pseudo-labels to predict the positive probability for each instance. The prediction results are then aggregated to obtain the bag-level prediction. Bag-based methods, on the other hand, utilize an instance-level feature extractor to extract the features of each instance within a bag. These features are then aggregated using a pooling function to obtain a bag feature. Subsequently, a bag-level classifier is trained to complete the bag-level prediction. Attention-based methods serve as the mainstream technique for feature aggregation \cite{8,9,10,11,12,13,14,37}, wherein attention scores for each instance are employed for instance-level classification. All these methods rely solely on labeled bags for model training, and their effectiveness heavily depends on the availability of a large amount of labeled bags. In contrast, we introduce the concept of weakly semi-supervised WSI classification in this paper and propose the CroCo framework to efficiently address this problem.

\subsection{Semi-supervised Classification.}

Semi-supervised Learning has been widely explored. Semi-supervised image classification studies can mainly be divided into pseudo-labeling methods \cite{55,56,57,81} and consistency regularization methods \cite{59,60,67,77,78,79}. Pseudo-labeling methods use a model pre-trained with labeled data to assign pseudo-labels to unlabeled data for subsequent training. Its main focus is on how to generate high-quality pseudo-labels while reducing the impact of pseudo-label noise. The consistency regulation methods follow the semi-supervised smoothness assumption \cite{66,80}, and achieves efficient use of unlabeled data by promoting consistent output of the same samples under different perturbations as constraints, such as Mean Teacher \cite{60}, UDA \cite{59}, and other methods. These methods mainly focus on the construction of diverse perturbations. The latest methods \cite{58,61,62,63,64,65} often combine the above two ideas. For example, FixMatch \cite{58} applies consistency regularization on given instances under weak and strong augmentations and filtering reliable pseudo-labels under threshold selection for weak augmentations. However, the WSWC task is different from common semi-supervised paradigm. In WSWC, only a subset of bags is labeled, while the labels of instances within positive bags remain unknown. Several studies have addressed semi-supervised WSI classification or MIL \cite{72,73,74,75,76}, but they all rely on instance-level annotations, which are different from WSWC. Therefore, there is currently no semi-supervised method that can effectively solve the WSWC task. 

\section{Preliminaries}

\subsection{Problem Formulation}
Given a dataset $W$ containing $N$ WSIs, and each WSI $W_i$ is divided into non-overlapping small patches $\{p_{i,j},j=1,2,\ldots,n_i\}$, where $n_i$ is the number of patches cut from $W_i$. All the patches in $W_i$ form a bag, where each patch is an instance of this bag. The bag label $Y_i\in\left\{0,1\right\},\ i=\{1,2,\ldots,N\}$ and the instance labels $\{y_{i,j},j=1,2,\ldots,n_i\}$ have the following relationship: 
\begin{equation}
  Y_i=\left\{\begin{array}{cc}
     0,& \quad \text { if } \sum_j y_{i, j}=0 \\
     1,& \quad \text { else }
  \end{array}\right.
  \label{eq1}
\end{equation}

This indicates that all instances in negative bags are negative, while at least one positive instance exists in a positive bag. In the weakly supervised MIL setting, only the labels of bags in the training set are available, while the labels of instances are unknown.

In the Weakly Semi-supervised WSI Classification (WSWC) task, the training dataset $D_{N+M}=D_N^l\cup D_M^u$ contains only a small proportion of labeled WSI $(W_i,Y_i)\in D_N^l,i=\{1,2,\ldots,N\}$, while the remaining $M$ WSI $W_i\in D_M^u,i=\{1,2,\ldots,M\}$ is unlabeled. Only the labels of instances in labeled negative WSIs are known to be negative, while the labels of all other instances are unknown. The goal of the WSWC task is to accurately predict the labels of each bag (bag classification) and each instance (instance classification) in the test set. 

\subsection{Bag-based MIL Method Using Attention for Feature Aggregation}
\label{sec32}

Attention is currently the mainstream feature aggregation methods in bag-based MIL methods, which usually consists of an instance-level encoder $f_{bag}$, an attention module $A_{bag}$, and a bag prediciton head $\varphi_{bag}$. First, $f_{bag}$ is used to extract features $z_{i,j}$ from all instances $\{p_{i,j},j=1,2,\ldots,n_i\}$ in bag $W_i$, and then these instance features are aggregated through an attention-based aggregation function to obtain the bag feature $Z_i$:
\begin{equation}
  Z_{i}=\sum_{j=1}^{n_{i}} a_{i, j} z_{i,j}, z_{i,j}= f_{bag}\left(p_{i, j}\right)
  \label{eq2}
\end{equation}
\begin{equation}
  a_{i,j}=\frac{\exp \left\{w^{\top} \tanh \left(V z_{i, j}^{\top}\right)\right\}}{\sum_{j=1}^{n_{i}} \exp \left\{w^{\top} \tanh \left(V z_{i, j}^{\top}\right)\right\}}
  \label{eq3}
\end{equation}

where $a_{i,j}$ is the attention score predicted by the self-attention network parameterized by $w$ and $V$. The weight $a_{i,j}$ is a score for each instance, which essentially reflects the probability that the instance is positive. Therefore, it can also be normalized and used as the prediction result for each instance:
\begin{equation}
  \hat{y}_{i, j}=\operatorname{norm}\left(a_{i, j}\right)
  \label{eq4}
\end{equation}
where $\hat{y}_{i,j}$ represents the prediction for each instance, and $norm(\cdot)$ is the normalization function.

Finally, the bag prediction head $\varphi_{bag}$ is used to predict the class of the bag:
\begin{equation}
  \widehat{Y}_{i}=\varphi_{bag}\left(Z_{i}\right)
  \label{eq5}
\end{equation}

Since the bag label $Y_i$ is known, the bag-based classifier can be trained end-to-end with the following Cross Entropy loss.
\begin{equation}
  \text {Loss}=C E\left(Y_{i}, \widehat{Y}_{i}\right)
  \label{eq6}
\end{equation}

\subsection{Instance-based MIL Method}
\label{sec33}

Instance-based MIL methods trains an instance classifier with pseudo-labels to predict the probability of being positive for each instance, and then aggregates the predictions to obtain the bag prediction. The instance-level classifier generally consists of an instance-level encoder $f_{ins}$ and an instance prediction head $\varphi_{ins}$. For instances from negative bags, their labels are 0; for instances from positive bags, multiple methods can be explored to assign them pseudo-labels $\tilde{y}_{i,j}$ \cite{1,2,3}. The loss function of the instance-level classifier is the cross-entropy between the network prediction $\hat{y}_{i,j}$ and the pseudo-label $\tilde{y}_{i,j}$. 

\begin{equation}
\begin{gathered}
\hat{y}_{i, j}=\varphi_{i n s}\left(z_{i, j}\right), z_{i, j}=f_{ins}\left(p_{i, j}\right) \\
y_{i, j}=\left\{\begin{array}{cc}
\tilde{y}_{i, j}, & \text { if } Y_i=1 \\
0, & \text { else }
\end{array}\right. \\
\text { Loss }=C E\left(y_{i, j}, \hat{y}_{i, j}\right)
\end{gathered}
\label{eq7}
\end{equation}

Bag prediction can then be obtained by aggregating instance predictions by mean pooling, max pooling or other pooling methods.

\begin{figure*}[t!]
  \centering
  \includegraphics[width=\textwidth]{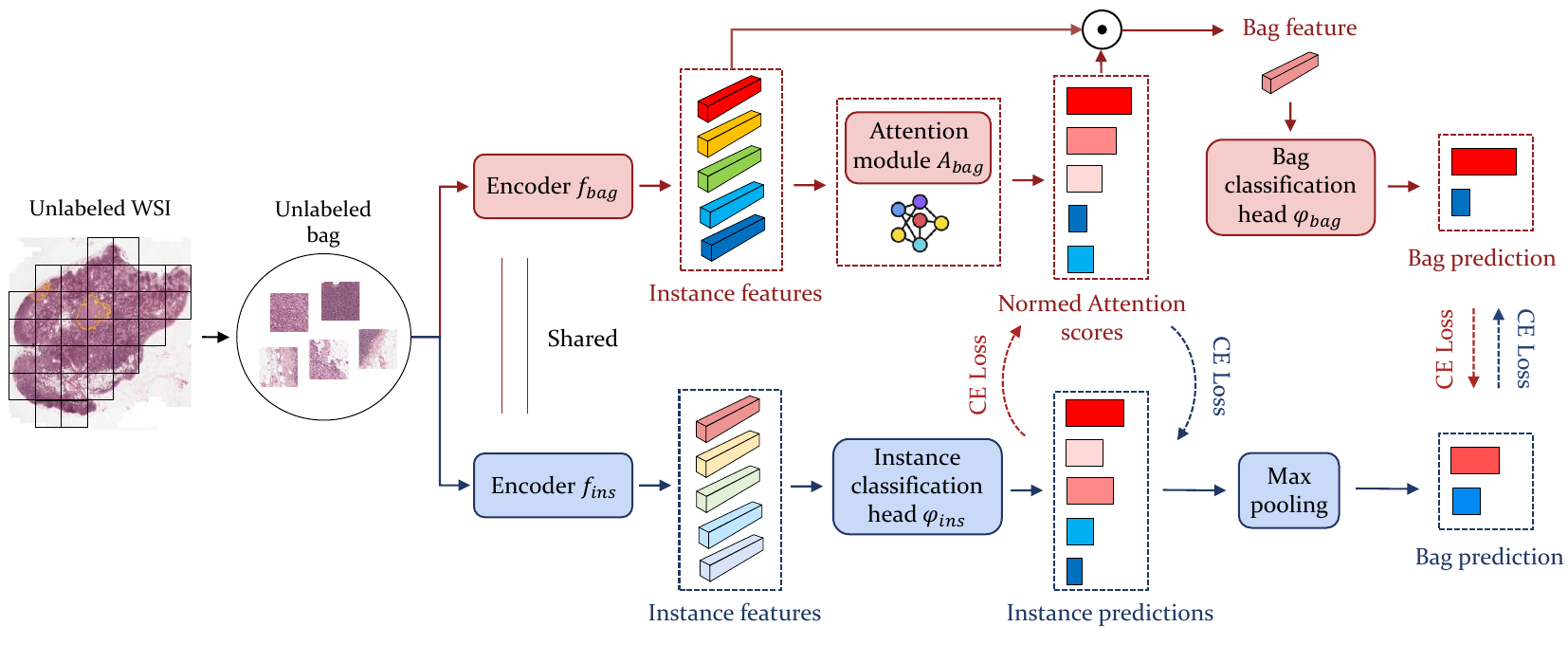}
  \caption{Pipeline of our Framework CroCo in processing unlabeled bags.}
  \label{figure2}
\end{figure*}

\subsection{Our Method}
As shown in Figure \ref{figure2}, the proposed CroCo is a dual-branch framework, where the upper branch is an bag-based classifier using attention for feature aggregation (see Section \ref{sec32}), and the lower branch is an instance-based classifier (see Section \ref{sec33}). The two branches share the same instance encoder. The main idea of CroCo is to establish both bag-level and instance-level cross-consistency supervision between the two branches. Algorithm 1 in the Supplementary Materials presents the pseudocode and loss function of CroCo. CroCo is trained as follows.

\textbf{For bags with labels}, both branches use their true labels for bag-level supervision. Meanwhile, for instances from positive bags, the normalized attention scores produced by the attention module of the bag-based classifier are employed as pseudo-labels for the instance classifier. Simultaneously, the predictions of the instance-based classifier serve as pseudo-labels for each instance within the attention module, thereby establishing cross-supervision at the instance-level. Conversely, for instances from negative bags, cross-supervision is omitted, and the instance-based classifier is directly trained using the true negative instance labels.

\textbf{For unlabeled bags}, both branches employ cross-consistency supervision at both the bag level and the instance level. At the bag level, the upper branch's predicted bag scores are employed as pseudo-labels for the lower branch, and vice versa. Similarly, at the instance level, the attention module of the upper branch generates normalized attention scores, which are utilized as pseudo-labels for the instance-based classifier. Concurrently, the instance-based classifier's predictions from the lower branch serve as pseudo-labels for the attention module in the upper branch.

During inference we use the prediction results of the bag classification head $\varphi_{bag}$ as the bag prediction results and use the prediction results of the instance classification head $\varphi_{ins}$ as the instance prediction results.

Algorithm \ref{alg1} presents the pseudocode and loss function of CroCo.

\begin{algorithm}[t!]
  \caption{Pseudo-code of CroCo in a PyTorch-like style.}
  \label{alg1}
  \textbf{Input}: Current Bag $W_i$ and its true label $Y_i$ (optional), attention-based bag-level classifier: Encoder $f_{baq}$, Attention Module $A_{bag}$ and Bag classification head $\varphi_{bag}$; instance-level classifier: Encoder $f_{ins}$ and Instance classification head $\varphi_{ins}$.\\
  \textbf{Output}: predictions of $\varphi_{bag}$ and $\varphi_{ins}$.\\
  \begin{algorithmic}[1] 
  \State \# Training process
  \For{$W_i$ in loader}
  \State \# Forward the attention-based bag-level classifier
  \State $\begin{aligned}
  & z_{i, j}^{bag}=f_{bag}\left(p_{i, j}\right) \\
  & \hat{y}_{i, j}^{bag}=\text{norm}\left(a_{i, j}^{bag}\right),\\
  & a_{i, j}^{bag}=A_{bag}\left(z_{i, j}^{bag} \mid z_{i, 1}^{bag}, z_{i, 2}^{bag}, \ldots, z_{i, n_i}^{bag}\right) \\
  & \hat{Y}_i^{bag}=\varphi_{bag}\left(Z_i^{bag}\right), Z_i^{bag}=\sum_{j=1}^{n_i} a_{i, j}^{bag} z_{i, j}^{bag}
  \end{aligned}$
  \State \# Forward the instance-level classifier
  \State $\begin{aligned}
  & \hat{y}_{i, j}^{ins}=\varphi_{ins}\left(z_{i, j}^{ins}\right), z_{i, j}^{ins}=f_{ins}\left(p_{i, j}\right) \\
  & \hat{Y}_i^{ins}=\text{Max-pooling}\left(\hat{y}_{i, j}^{ins}\right)
  \end{aligned}$
  \State \# Calculate instance level loss and bag level loss
  \If{$W_i$ has true label}
  \State $\text{Loss}_{bag\_level} = CE\left(Y_i, \hat{Y}_i^{bag}\right) + CE\left(Y_i, \hat{Y}_i^{ins}\right)$
  \If{$Y_i==0$}
  \State $\text{Loss}_{ins\_level} = CE\left(0, \hat{y}_{i, j}^{ins}\right)$
  \Else
  \State $\text{Loss}_{ins\_level} = CE\left(\hat{y}_{i, j}^{bag}, \hat{y}_{i, j}^{ins}\right) + CE\left(\hat{y}_{i, j}^{ins}, \hat{y}_{i, j}^{bag}\right)$
  \EndIf
  \State $\text{Loss}_{bag\_level} = CE\left(\hat{Y}_i^{ins}, \hat{Y}_i^{bag}\right) + CE\left(\hat{Y}_i^{bag}, \hat{Y}_i^{ins}\right)$
  \EndIf
  \EndFor
  \end{algorithmic}
\end{algorithm}
\begin{algorithm}[t]
\ContinuedFloat
  \caption{Pseudo-code of CroCo in a PyTorch-like style.}
  \begin{algorithmic}[1] 
  \State \# Calculate loss for labeled sample and unlabeled sample including two-level losses
  \State $ \begin{aligned}
  & \text{Loss}_{\text{sup}} = \frac{1}{N} \sum_{W_i, Y_i \in D_N^l} \left(\text{Loss}_{\text{ins\_level}} + \text{Loss}_{\text{bag\_level}}\right), \\
  & \text{Loss}_{\text{unsup}} = \frac{1}{M} \sum_{W_i \in D_M^u} \left(\text{Loss}_{\text{ins\_level}} + \text{Loss}_{\text{bag\_level}}\right)
  \end{aligned} $
  \State \# Calculate total loss including two-level losses
  \State $ \begin{aligned}
  \text{Total\_Loss} = \text{Loss}_{\text{sup}} + \tau \cdot \text{Loss}_{\text{unsup}}
  \end{aligned} $
  \State \# Testing process
  \State Make bag-level predictions using $\varphi_{bag}$.
  \State Make instance-level predictions using $\varphi_{ins}$.
  \end{algorithmic}
  \label{alg:continued}
\end{algorithm}

\section{Experiment}

\subsection{Datasets}

We used four datasets to comprehensively evaluate the instance classification and bag classification performance of CroCo, including one synthetic dataset and three real-world datasets. In order to explore the performance of CroCo under different positive instance ratios and labeled data ratios, we used a 10-category natural image dataset CIFAR 10 \cite{49} to construct WSI datasets (referred to as CIFAR-10-MIL dataset) under different conditions. In addition, we used three real-world datasets from different centers and different cancers to evaluate the performance of CroCo in real pathological diagnosis under different labeled data ratios. The real-world datasets include a breast cancer lymph node metastasis public dataset Camelyon16 Dataset \cite{32}, a lung cancer diagnosis public dataset TCGA Lung Cancer Dataset\footnote {http://www.cancer.gov/tcga}, and a cervical cancer lymph node metastasis in-house dataset Clinical Cervical Dataset. A detailed description of the datasets is as follows.

\subsubsection{Synthetic CIFAR-10-MIL Dataset}

In order to assess the performance of CroCo, a comparison was made under various positive instance ratios and labeled data ratios. Similar to WENO \cite{18}, the CIFAR-10 dataset \cite{49}, consisting of natural images across ten classes, was employed to generate a synthetic Whole Slide Image (WSI) dataset named CIFAR-10-MIL Dataset. This dataset was created with varying positive ratios.

The CIFAR-10 dataset comprises a total of 60,000 color images, each measuring 32$\times$32 pixels. These images are categorized into ten classes, namely airplane, automobile, bird, cat, deer, dog, frog, horse, ship, and truck, with each category containing 6,000 images. For training and testing purposes, 50,000 and 10,000 images, respectively, were utilized.

To simulate WSIs with pathological characteristics, we combined a random selection of images from each category of the CIFAR-10 dataset. Specifically, we treated each image within a category as an instance, designating only the instances from the "truck" category as positive, while labeling the remaining instances as negative (the choice of the truck category was random).

Next, we formed positive bags by randomly selecting $a$ positive instances and $100-a$ negative instances (without repetition) from all available instances. Additionally, we created negative bags consisting of 100 negative instances (without repetition). This process was repeated until all positive and negative instances from the CIFAR-10 dataset were exhausted. By adjusting the value of '$a$', we generated six subsets of the CIFAR-10-MIL dataset with positive ratios of 1\%, 5\%, 10\%, 20\%, 50\%, and 70\%, respectively.
  
\subsubsection{Camelyon16 Dataset}
The Camelyon16 dataset, which is publicly available, serves as a resource for detecting breast cancer metastasis in lymph nodes \cite{32}. It encompasses 400 H\&E-stained WSIs of lymph nodes, with 270 allocated for training and 130 for testing. WSIs containing metastasis are designated as positive, while those without metastasis are classified as negative. Notably, the dataset offers both slide-level labels denoting the positivity or negativity of a WSI and pixel-level labels identifying specific areas of metastasis.

To accommodate weakly supervised scenarios, our approach for training solely employed slide-level labels. For evaluating the instance classification performance of each algorithm, we utilized the pixel-level labels for cancerous regions. Similar to the preprocessing in \cite{13,18}, before initiating the training process, we divided each WSI into non-overlapping image patches measuring 512$\times$512 pixels at 10$\times$ magnification. Patches with an entropy value lower than 5 were eliminated as background noise. Moreover, a patch was deemed positive if it encompassed 25\% or more cancerous areas; otherwise, it received a negative label. Consequently, we obtained a total of 186,604 instances for analysis.

\subsubsection{TCGA Lung Cancer Dataset}

The TCGA Lung Cancer dataset, available from the Cancer Genome Atlas (TCGA) Data Portal, encompasses a total of 1054 Whole Slide Images (WSIs). This dataset focuses on two distinct subtypes of lung cancer, namely Lung Adenocarcinoma and Lung Squamous Cell Carcinoma. Our primary objective is to achieve an accurate diagnosis for both subtypes, wherein WSIs depicting Lung Adenocarcinoma are labeled as negative, while WSIs featuring Lung Squamous Cell Carcinoma are labeled as positive.

It is important to note that this dataset provides solely slide-level labels, and patch-level labels are not available. After preprocessing similar to \cite{13,18}, at a magnification of 20$\times$, the dataset comprises approximately 5.2 million patches, with an average of around 5,000 patches per slide. To facilitate the partitioning of the dataset, 840 slides were assigned for training purposes, while 210 slides were designated for testing. It is worth mentioning that four slides exhibiting low quality or corruption were excluded from the dataset.

\subsubsection{Clinical Cervical Dataset}
The Clinical Cervical dataset utilized in this study is an in-house clinical pathology dataset, comprising 374 H\&E-stained WSIs of primary cervical cancer lesions obtained from diverse patients, following a careful slide selection process. All patients included in the dataset underwent abdominal hysterectomy, along with pelvic lymph node dissection ± para-aortic lymph node dissection. The lymph node status of each patient was meticulously determined by professional gynecological pathologists subsequent to surgery. Importantly, all patients have comprehensive follow-up records spanning over a five-year period, along with results from significant immunohistochemical marker detections.

Our aim was to employ this dataset to address the challenge of predicting the prognosis of patient survival, a task that cannot be directly determined from the H\&E slides by medical professionals. By utilizing deep learning technology, we sought to predict this clinical outcome directly from the H\&E-stained slides, which holds crucial clinical significance. Deep learning technology has the capability to extract extensive clinical information from H\&E-stained slides and accurately forecast the survival of patients with cervical cancer. Such predictions serve as vital evidence for selecting appropriate treatment plans and evaluating treatment effectiveness.

It is noteworthy that in clinical practice, acquiring prognosis labels often necessitates long-term and meticulous follow-up, thereby presenting significant challenges. In this context, CroCo proves to be a valuable asset by effectively reducing the labeling requirements by half. This reduction offers valuable assistance in overcoming the challenges associated with acquiring prognosis labels and contributes to the advancement of research in this field.

Similar to the preprocessing in \cite{18}, the experiment was conducted using a magnification of 5$\times$, and each WSI was divided into non-overlapping patches measuring 224$\times$224 pixels, which were subsequently assembled into bags. Patches deemed background, with entropy values below 5, were eliminated from the original WSI. To follow the approach outlined by Skrede et al. \cite{84}, we categorized all patients based on comprehensive follow-up records, utilizing the median as a cutoff. Patients who did not experience cancer-related death within a three-year timeframe were labeled as negative (indicating a good prognosis), while those who did were labeled as positive (indicating a poor prognosis). Subsequently, the WSIs were randomly divided into a training set consisting of 294 cases and a test set consisting of 80 cases, based on these labels.

\subsection{Evaluation Metrics}
We used AUC as the evaluation metric for both bag classification and instance classification tasks. We report the AUCs for bag classification and instance classification on the CIFAR-10-MIL dataset and the Camelyon 16 dataset, respectively. However, due to the unavailability of ground-truth instance labels for the TCGA Lung Cancer dataset and the Clinical Cervical dataset, we solely report the AUC scores for bag classification on these datasets.

\subsection{Comparison Methods}
The model trained with only labeled bags but without unlabeled bags are considered as baseline. No existing semi-supervised methods can be directly used in the WSWC problem, so we adapt several classical and latest semi-supervised leaning methods in the natural image processing field, including Mean Teacher \cite{60}, FixMatch \cite{58} and FreeMatch \cite{64} for comparison. We used an attention-based bag-level classifier as the backbone for all comparison networks, and implemented the semi-supervised strategy in bag-level as in the original method. Specifically, in these comparison methods, weak and strong image augmentations at the patch-level were replaced with random noise addition and random dropout at the bag-level. For real-world pathological datasets, these operations were applied at the feature level.

\subsection{Implementation Details}
We evaluated CroCo using two advanced bag-based MIL methods, ABMIL \cite{8} and DSMIL \cite{13}, as the backbone of the bag-based classifier branch. We followed WENO \cite{18} and DSMIL \cite{13} for the preprocessing of the CIFAR-10-MIL dataset and the other three real-world datasets. 
Cross Entropy loss is used for model training. The SGD optimizer with an initial learning rate of 0.001 is used for parameter updating during training. All methods are implemented in the PyTorch framework, and experiments are conducted using four 3090 GPUs. The optimal hyperparameters vary depending on the datasets, and we use grid search to determine the optimal values on the validation set. 

\begin{table*}[t!]
\caption{The performance of bag-level and instance-level classification on the CIFAR-MIL Dataset. The bag-based classifier branch used ABMIL \cite{8}. Full-data represents training ABMIL using 100\% labeled data ratio.}
\begin{adjustbox}{width=\textwidth}
\begin{tabular}{cc|cccccc|cccccc}
\hline
\multicolumn{2}{c|}{Backbone:ABMIL}                                                          & \multicolumn{6}{c|}{Performance   of bag-level classification}                                                                                                                                                                                                                                                     & \multicolumn{6}{c}{Performance of   instance-level classification}                                                                                                                                                                                                                                                 \\ \hline
                                          &                          & \multicolumn{6}{c|}{labeled data ratio}                                                                                                                                                                                                                                                                                 & \multicolumn{6}{c}{labeled data ratio}                                                                                                                                                                                                                                                                                    \\ \cline{3-14} 
\multirow{-2}{*}{positive instance ratio} & \multirow{-2}{*}{Method} & 10\%                                   & 20\%                                   & 30\%                                   & 40\%                                   & \multicolumn{1}{c|}{50\%}                                                                                & \multicolumn{1}{c|}{Full-data}                                                                        & 10\%                                   & 20\%                                   & 30\%                                   & 40\%                                   & \multicolumn{1}{c|}{50\%}                                                                                & Full-data                                                                        \\ \hline
                                          & Baseline                                         & 0.4686                                 & 0.5170                                 & 0.5336                                 & 0.5440                                 & \multicolumn{1}{c|}{0.5580}                                 & {\color[HTML]{808080} }                                  & 0.6330                                 & 0.6768                                 & 0.7025                                 & 0.7339                                 & \multicolumn{1}{c|}{0.7439}                                 & {\color[HTML]{808080} }                                  \\
                                          & MT   (17'NeurIPS)                                & 0.5417                                 & 0.5709                                 & 0.6119                                 & 0.6273                                 & \multicolumn{1}{c|}{0.6363}                                 & {\color[HTML]{808080} }                                  & 0.6470                                 & 0.6891                                 & 0.7090                                 & 0.7370                                 & \multicolumn{1}{c|}{0.7510}                                 & {\color[HTML]{808080} }                                  \\
                                          & Fixmatch   (20'NeurIPS)                          & 0.6002                                 & 0.6247                                 & 0.6332                                 & 0.6425                                 & \multicolumn{1}{c|}{0.6453}                                 & {\color[HTML]{808080} }                                  & 0.6561                                 & 0.6933                                 & 0.7134                                 & 0.7366                                 & \multicolumn{1}{c|}{0.7535}                                 & {\color[HTML]{808080} }                                  \\
                                          & Freematch   (22'ICLR)                            & 0.6296                                 & 0.6412                                 & 0.6458                                 & 0.6519                                 & \multicolumn{1}{c|}{0.6587}                                 & {\color[HTML]{808080} }                                  & 0.6966                                 & 0.7041                                 & 0.7119                                 & 0.7461                                 & \multicolumn{1}{c|}{0.7551}                                 & {\color[HTML]{808080} }                                  \\
\multirow{-5}{*}{1\%}                     & \textbf{CroCo   (Ours)}                           & {\color[HTML]{FF0000} \textbf{0.6454}} & {\color[HTML]{FF0000} \textbf{0.6533}} & {\color[HTML]{FF0000} \textbf{0.6617}} & {\color[HTML]{FF0000} \textbf{0.6696}} & \multicolumn{1}{c|}{{\color[HTML]{FF0000} \textbf{0.6706}}} & \multirow{-5}{*}{{\color[HTML]{808080} \textbf{0.6883}}} & {\color[HTML]{FF0000} \textbf{0.7160}} & {\color[HTML]{FF0000} \textbf{0.7289}} & {\color[HTML]{FF0000} \textbf{0.7331}} & {\color[HTML]{FF0000} \textbf{0.7650}} & \multicolumn{1}{c|}{{\color[HTML]{FF0000} \textbf{0.7718}}} & \multirow{-5}{*}{{\color[HTML]{808080} \textbf{0.7553}}} \\ \hline
                                          & Baseline                                         & 0.6121                                 & 0.6372                                 & 0.6807                                 & 0.7371                                 & \multicolumn{1}{c|}{0.7643}                                 & {\color[HTML]{808080} }                                  & 0.7519                                 & 0.8114                                 & 0.8208                                 & 0.8573                                 & \multicolumn{1}{c|}{0.8624}                                 & {\color[HTML]{808080} }                                  \\
                                          & MT   (17'NeurIPS)                                & 0.6474                                 & 0.6849                                 & 0.7078                                 & 0.7564                                 & \multicolumn{1}{c|}{0.7889}                                 & {\color[HTML]{808080} }                                  & 0.7678                                 & 0.8249                                 & 0.8469                                 & 0.8627                                 & \multicolumn{1}{c|}{0.8717}                                 & {\color[HTML]{808080} }                                  \\
                                          & Fixmatch   (20'NeurIPS)                          & 0.6436                                 & 0.6898                                 & 0.7164                                 & 0.7515                                 & \multicolumn{1}{c|}{0.7891}                                 & {\color[HTML]{808080} }                                  & 0.7613                                 & 0.8221                                 & 0.8434                                 & 0.8613                                 & \multicolumn{1}{c|}{0.8726}                                 & {\color[HTML]{808080} }                                  \\
                                          & Freematch   (22'ICLR)                            & 0.6874                                 & 0.7249                                 & 0.7478                                 & 0.7764                                 & \multicolumn{1}{c|}{0.8027}                                 & {\color[HTML]{808080} }                                  & 0.7657                                 & 0.8234                                 & 0.8445                                 & 0.8600                                 & \multicolumn{1}{c|}{0.8735}                                 & {\color[HTML]{808080} }                                  \\
\multirow{-5}{*}{5\%}                     & \textbf{CroCo   (Ours)}                           & {\color[HTML]{FF0000} \textbf{0.7675}} & {\color[HTML]{FF0000} \textbf{0.7744}} & {\color[HTML]{FF0000} \textbf{0.7928}} & {\color[HTML]{FF0000} \textbf{0.8011}} & \multicolumn{1}{c|}{{\color[HTML]{FF0000} \textbf{0.8133}}} & \multirow{-5}{*}{{\color[HTML]{808080} \textbf{0.8850}}} & {\color[HTML]{FF0000} \textbf{0.7998}} & {\color[HTML]{FF0000} \textbf{0.8362}} & {\color[HTML]{FF0000}  \textbf{0.8559}} & {\color[HTML]{FF0000} \textbf{0.8713}} & \multicolumn{1}{c|}{{\color[HTML]{FF0000} \textbf{0.8871}}} & \multirow{-5}{*}{{\color[HTML]{808080} \textbf{0.9083}}} \\ \hline
                                          & Baseline                                         & 0.7612                                 & 0.7723                                 & 0.7882                                 & 0.8143                                 & \multicolumn{1}{c|}{0.8478}                                 & {\color[HTML]{808080} }                                  & 0.8224                                 & 0.8454                                 & 0.8570                                 & 0.8726                                 & \multicolumn{1}{c|}{0.8857}                                 & {\color[HTML]{808080} }                                  \\
                                          & MT   (17'NeurIPS)                                & 0.7855                                 & 0.8208                                 & 0.8307                                 & 0.8515                                 & \multicolumn{1}{c|}{0.8610}                                 & {\color[HTML]{808080} }                                  & 0.8356                                 & 0.8687                                 & 0.8770                                 & 0.8891                                 & \multicolumn{1}{c|}{0.8961}                                 & {\color[HTML]{808080} }                                  \\
                                          & Fixmatch   (20'NeurIPS)                          & 0.7884                                 & 0.8212                                 & 0.8399                                 & 0.8503                                 & \multicolumn{1}{c|}{0.8598}                                 & {\color[HTML]{808080} }                                  & 0.8369                                 & 0.8623                                 & 0.8767                                 & 0.8895                                 & \multicolumn{1}{c|}{0.8918}                                 & {\color[HTML]{808080} }                                  \\
                                          & Freematch   (22'ICLR)                            & 0.7990                                 & 0.8288                                 & 0.8441                                 & 0.8646                                 & \multicolumn{1}{c|}{0.8696}                                 & {\color[HTML]{808080} }                                  & 0.8370                                 & 0.8518                                 & 0.8641                                 & 0.8848                                 & \multicolumn{1}{c|}{0.8933}                                 & {\color[HTML]{808080} }                                  \\
\multirow{-5}{*}{10\%}                    & \textbf{CroCo   (Ours)}                           & {\color[HTML]{FF0000} \textbf{0.8579}} & {\color[HTML]{FF0000}  \textbf{0.8693}} & {\color[HTML]{FF0000} \textbf{0.8771}} & {\color[HTML]{FF0000} \textbf{0.8850}} & \multicolumn{1}{c|}{{\color[HTML]{FF0000} \textbf{0.9087}}} & \multirow{-5}{*}{{\color[HTML]{808080} \textbf{0.9955}}} & {\color[HTML]{FF0000} \textbf{0.8424}} & {\color[HTML]{FF0000} \textbf{0.8854}} & {\color[HTML]{FF0000} \textbf{0.8876}} & {\color[HTML]{FF0000} \textbf{0.8956}} & \multicolumn{1}{c|}{{\color[HTML]{FF0000} \textbf{0.9057}}} & \multirow{-5}{*}{{\color[HTML]{808080} \textbf{0.9241}}} \\ \hline
                                          & Baseline                                         & 0.9485                                 & 0.9536                                 & 0.9650                                 & 0.9688                                 & \multicolumn{1}{c|}{0.9772}                                 & {\color[HTML]{808080} }                                  & 0.8357                                 & 0.8486                                 & 0.8645                                 & 0.8802                                 & \multicolumn{1}{c|}{0.8987}                                 & {\color[HTML]{808080} }                                  \\
                                          & MT   (17'NeurIPS)                                & 0.9636                                 & 0.9756                                 & 0.9905                                 & 0.9941                                 & \multicolumn{1}{c|}{0.9986}                                 & {\color[HTML]{808080} }                                  & 0.8495                                 & 0.8547                                 & 0.8721                                 & 0.8899                                 & \multicolumn{1}{c|}{0.9004}                                 & {\color[HTML]{808080} }                                  \\
                                          & Fixmatch   (20'NeurIPS)                          & 0.9760                                 & 0.9832                                 & 0.9918                                 & 0.9932                                 & \multicolumn{1}{c|}{0.9984}                                 & {\color[HTML]{808080} }                                  & 0.8408                                 & 0.8536                                 & 0.8747                                 & 0.8976                                 & \multicolumn{1}{c|}{0.9015}                                 & {\color[HTML]{808080} }                                  \\
                                          & Freematch   (22'ICLR)                            & 0.9820                                 & 0.9856                                 & 0.9912                                 & 0.9983                                 & \multicolumn{1}{c|}{0.9986}                                 & {\color[HTML]{808080} }                                  & 0.8454                                 & 0.8654                                 & 0.8823                                 & 0.8998                                 & \multicolumn{1}{c|}{0.9023}                                 & {\color[HTML]{808080} }                                  \\
\multirow{-5}{*}{20\%}                    & \textbf{CroCo   (Ours)}                           & {\color[HTML]{FF0000} \textbf{0.9972}} & {\color[HTML]{FF0000} \textbf{1.0000}} & {\color[HTML]{FF0000} \textbf{1.0000}} & {\color[HTML]{FF0000} \textbf{1.0000}} & \multicolumn{1}{c|}{{\color[HTML]{FF0000} \textbf{1.0000}}} & \multirow{-5}{*}{{\color[HTML]{808080} \textbf{1.0000}}} & {\color[HTML]{FF0000} \textbf{0.8757}} & {\color[HTML]{FF0000} \textbf{0.8892}} & {\color[HTML]{FF0000} \textbf{0.8900}} & {\color[HTML]{FF0000} \textbf{0.9034}} & \multicolumn{1}{c|}{{\color[HTML]{FF0000} \textbf{0.9195}}} & \multirow{-5}{*}{{\color[HTML]{808080} \textbf{0.9237}}} \\ \hline
                                          & Baseline                                         & 1.0000                                 & 1.0000                                 & 1.0000                                 & 1.0000                                 & \multicolumn{1}{c|}{1.0000}                                 & {\color[HTML]{808080} }                                  & 0.7630                                 & 0.7721                                 & 0.7989                                 & 0.8046                                 & \multicolumn{1}{c|}{0.8189}                                 & {\color[HTML]{808080} }                                  \\
                                          & MT   (17'NeurIPS)                                & 1.0000                                 & 1.0000                                 & 1.0000                                 & 1.0000                                 & \multicolumn{1}{c|}{1.0000}                                 & {\color[HTML]{808080} }                                  & 0.8455                                 & 0.8503                                 & 0.8619                                 & 0.8702                                 & \multicolumn{1}{c|}{0.8787}                                 & {\color[HTML]{808080} }                                  \\
                                          & Fixmatch   (20'NeurIPS)                          & 1.0000                                 & 1.0000                                 & 1.0000                                 & 1.0000                                 & \multicolumn{1}{c|}{1.0000}                                 & {\color[HTML]{808080} }                                  & 0.8495                                 & 0.8532                                 & 0.8617                                 & 0.8713                                 & \multicolumn{1}{c|}{0.8798}                                 & {\color[HTML]{808080} }                                  \\
                                          & Freematch   (22'ICLR)                            & 1.0000                                 & 1.0000                                 & 1.0000                                 & 1.0000                                 & \multicolumn{1}{c|}{1.0000}                                 & {\color[HTML]{808080} }                                  & 0.8449                                 & 0.8518                                 & 0.8633                                 & 0.8754                                 & \multicolumn{1}{c|}{0.8818}                                 & {\color[HTML]{808080} }                                  \\
\multirow{-5}{*}{50\%}                    & \textbf{CroCo   (Ours)}                           & 1.0000                                 & 1.0000                                 & 1.0000                                 & 1.0000                                 & \multicolumn{1}{c|}{1.0000}                                 & \multirow{-5}{*}{{\color[HTML]{808080} \textbf{1.0000}}} & {\color[HTML]{FF0000} \textbf{0.8668}} & {\color[HTML]{FF0000} \textbf{0.8735}} & {\color[HTML]{FF0000} \textbf{0.8798}} & {\color[HTML]{FF0000} \textbf{0.8892}} & \multicolumn{1}{c|}{{\color[HTML]{FF0000} \textbf{0.9002}}} & \multirow{-5}{*}{{\color[HTML]{808080} \textbf{0.8224}}} \\ \hline
                                          & Baseline                                         & 1.0000                                 & 1.0000                                 & 1.0000                                 & 1.0000                                 & \multicolumn{1}{c|}{1.0000}                                 & {\color[HTML]{808080} }                                  & 0.7000                                 & 0.7224                                 & 0.7394                                 & 0.7517                                 & \multicolumn{1}{c|}{0.7743}                                 & {\color[HTML]{808080} }                                  \\
                                          & MT   (17'NeurIPS)                                & 1.0000                                 & 1.0000                                 & 1.0000                                 & 1.0000                                 & \multicolumn{1}{c|}{1.0000}                                 & {\color[HTML]{808080} }                                  & 0.8442                                 & 0.8489                                 & 0.8511                                 & 0.8663                                 & \multicolumn{1}{c|}{0.8769}                                 & {\color[HTML]{808080} }                                  \\
                                          & Fixmatch   (20'NeurIPS)                          & 1.0000                                 & 1.0000                                 & 1.0000                                 & 1.0000                                 & \multicolumn{1}{c|}{1.0000}                                 & {\color[HTML]{808080} }                                  & 0.8316                                 & 0.8400                                 & 0.8456                                 & 0.8569                                 & \multicolumn{1}{c|}{0.8637}                                 & {\color[HTML]{808080} }                                  \\
                                          & Freematch   (22'ICLR)                            & 1.0000                                 & 1.0000                                 & 1.0000                                 & 1.0000                                 & \multicolumn{1}{c|}{1.0000}                                 & {\color[HTML]{808080} }                                  & 0.8398                                 & 0.8480                                 & 0.8472                                 & 0.8635                                 & \multicolumn{1}{c|}{0.8740}                                 & {\color[HTML]{808080} }                                  \\
\multirow{-5}{*}{70\%}                    & \textbf{CroCo   (Ours)}                           & 1.0000                                 & 1.0000                                 & 1.0000                                 & 1.0000                                 & \multicolumn{1}{c|}{1.0000}                                 & \multirow{-5}{*}{{\color[HTML]{808080} \textbf{1.0000}}} & {\color[HTML]{FF0000} \textbf{0.8642}} & {\color[HTML]{FF0000} \textbf{0.8789}} & {\color[HTML]{FF0000} \textbf{0.8811}} & {\color[HTML]{FF0000} \textbf{0.8863}} & \multicolumn{1}{c|}{{\color[HTML]{FF0000} \textbf{0.8969}}} & \multirow{-5}{*}{{\color[HTML]{808080} \textbf{0.7935}}} \\ \hline
\end{tabular}
\label{table1}
\end{adjustbox}
\end{table*}

\begin{table*}[t!]
\caption{The performance of bag-level and instance-level classification on the Camelyon Dataset. Both ABMIL \cite{8} and DSMIL \cite{13} are used as the bag-based classifier. Full-data represents training the corresponding bag-based classifier using 100\% labeled data ratio.}
\begin{adjustbox}{width=\textwidth}
\begin{tabular}{c|cccccc|cccccc}
\hline
Backbone:ABMIL          & \multicolumn{6}{c|}{Performance of bag-level classification}                                                                                                                                                                                                                                                       & \multicolumn{6}{c}{Performance   of instance-level classification}                                                                                                                                                                                                                                                 \\ \hline
labeled data ratio             & 10\%                                   & 20\%                                    & 30\%                                    & 40\%                                    & \multicolumn{1}{c|}{50\%}                                    & Full-data                                                                        & 10\%                                   & 20\%                                    & 30\%                                    & 40\%                                    & \multicolumn{1}{c|}{50\%}                                    & Full-data                                                                        \\ \hline
Baseline                & 0.4558                                 & 0.5223                                 & 0.5966                                 & 0.6568                                 & \multicolumn{1}{c|}{0.7129}                                 & {\color[HTML]{808080} }                                  & 0.5298                                 & 0.6252                                 & 0.6881                                 & 0.7391                                 & \multicolumn{1}{c|}{0.7728}                                 & {\color[HTML]{808080} }                                  \\
MT   (17'NeurIPS)       & 0.6106                                 & 0.7098                                 & 0.7318                                 & 0.7715                                 & \multicolumn{1}{c|}{0.8003}                                 & {\color[HTML]{808080} }                                  & 0.7863                                 & 0.8113                                 & 0.8392                                 & 0.8452                                 & \multicolumn{1}{c|}{0.8584}                                 & {\color[HTML]{808080} }                                  \\
Fixmatch   (20'NeurIPS) & 0.6687                                 & 0.7227                                 & 0.7558                                 & 0.7895                                 & \multicolumn{1}{c|}{0.8164}                                 & {\color[HTML]{808080} }                                  & 0.7957                                 & 0.8251                                 & 0.8394                                 & 0.8467                                 & \multicolumn{1}{c|}{0.8609}                                 & {\color[HTML]{808080} }                                  \\
Freematch   (22'ICLR)   & 0.6703                                 & 0.7258                                 & 0.7786                                 & 0.7954                                 & \multicolumn{1}{c|}{0.8162}                                 & {\color[HTML]{808080} }                                  & 0.8227                                 & 0.8382                                 & 0.8513                                 & 0.8417                                 & \multicolumn{1}{c|}{0.8664}                                 & {\color[HTML]{808080} }                                  \\ 
\textbf{CroCo   (Ours)}  & {\color[HTML]{FF0000} \textbf{0.7688}} & {\color[HTML]{FF0000} \textbf{0.7827}} & {\color[HTML]{FF0000} \textbf{0.8025}} & {\color[HTML]{FF0000} \textbf{0.8145}} & \multicolumn{1}{c|}{{\color[HTML]{FF0000} \textbf{0.8352}}} & \multirow{-5}{*}{{\color[HTML]{808080} \textbf{0.8379}}} & {\color[HTML]{FF0000} \textbf{0.8739}} & {\color[HTML]{FF0000} \textbf{0.8982}} & {\color[HTML]{FF0000} \textbf{0.9059}} & {\color[HTML]{FF0000} \textbf{0.9101}} & \multicolumn{1}{c|}{{\color[HTML]{FF0000} \textbf{0.9127}}} & \multirow{-5}{*}{{\color[HTML]{808080} \textbf{0.8480}}} \\ \hline
\end{tabular}
\label{table2}
\end{adjustbox}

\begin{adjustbox}{width=\textwidth}
\begin{tabular}{ccccccc|cccccc}
\hline
{\color[HTML]{000000} Backbone:DSMIL}        & \multicolumn{6}{c|}{Performance of bag-level classification}                                                                                                                                                                                                                               & \multicolumn{6}{c}{Performance   of instance-level classification}                                                                                                                                                                                                                         \\ \hline
\multicolumn{1}{c|}{labeled data ratio}             & 10\%                                    & 20\%                                    & 30\%                                    & 40\%                                    & \multicolumn{1}{c|}{50\%}                                    & Full-data                                                & 10\%                                    & 20\%                                    & 30\%                                    & 40\%                                    & \multicolumn{1}{c|}{50\%}                                    & Full-data                                                \\ \hline
\multicolumn{1}{l|}{Baseline}                & 0.4758                                 & 0.5387                                 & 0.6238                                 & 0.6730                                 & \multicolumn{1}{l|}{0.7230}                                 & {\color[HTML]{656565} }                                  & 0.5334                                 & 0.6465                                 & 0.6948                                 & 0.7431                                 & \multicolumn{1}{l|}{0.7825}                                 & {\color[HTML]{656565} }                                  \\
\multicolumn{1}{l|}{MT   (17'NeurIPS)}       & 0.7184                                 & 0.7535                                 & 0.7743                                 & 0.7987                                 & \multicolumn{1}{l|}{0.8188}                                 & {\color[HTML]{656565} }                                  & 0.7906                                 & 0.8223                                 & 0.8398                                 & 0.8513                                 & \multicolumn{1}{l|}{0.8719}                                 & {\color[HTML]{656565} }                                  \\
\multicolumn{1}{l|}{Fixmatch   (20'NeurIPS)} & 0.6748                                 & 0.7767                                 & 0.7813                                 & 0.8165                                 & \multicolumn{1}{l|}{0.8265}                                 & {\color[HTML]{656565} }                                  & 0.8090                                 & 0.8328                                 & 0.8414                                 & 0.8605                                 & \multicolumn{1}{l|}{0.8835}                                 & {\color[HTML]{656565} }                                  \\
\multicolumn{1}{l|}{Freematch   (22'ICLR)}   & 0.6827                                 & 0.7849                                 & 0.7965                                 & 0.8132                                 & \multicolumn{1}{l|}{0.8282}                                 & {\color[HTML]{656565} }                                  & 0.8119                                 & 0.8400                                 & 0.8454                                 & 0.8664                                 & \multicolumn{1}{l|}{0.8854}                                 & {\color[HTML]{656565} }                                  \\ 
\multicolumn{1}{l|}{\textbf{CroCo   (Ours)}}           & {\color[HTML]{FF0000} \textbf{0.7849}} & {\color[HTML]{FF0000} \textbf{0.8149}} & {\color[HTML]{FF0000} \textbf{0.8259}} & {\color[HTML]{FF0000} \textbf{0.8321}} & \multicolumn{1}{l|}{{\color[HTML]{FF0000} \textbf{0.8386}}} & \multirow{-5}{*}{{\color[HTML]{656565} \textbf{0.8401}}} & {\color[HTML]{FF0000} \textbf{0.8761}} & {\color[HTML]{FF0000} \textbf{0.8991}} & {\color[HTML]{FF0000} \textbf{0.9066}} & {\color[HTML]{FF0000} \textbf{0.9147}} & \multicolumn{1}{l|}{{\color[HTML]{FF0000} \textbf{0.9184}}} & \multirow{-5}{*}{{\color[HTML]{656565} \textbf{0.8568}}} \\ \hline
\end{tabular}
\end{adjustbox}
\end{table*}

\subsection{Results on the CIFAR-MIL Dataset}

We evaluated the performance of CroCo on the CIFAR-10-MIL dataset from two aspects: its performance under different positive instance ratios (PIRs) of the dataset and its performance under different labeled data ratios. The results are shown in Table \ref{table1}, where full-data represents labeling all bags to train the bag-based classifiers of CroCo. CroCo achieves the best bag and instance classification performance in all combination of PIRs and labeled ratios and significantly outperforms all compared methods.

One noteworthy phenomenon that warrants further investigation is the robust instance-level classification ability of CroCo. Interestingly, with a labeled data ratio of 50\%, CroCo demonstrates superior instance classification AUC compared to using the entire dataset with a labeled data ratio of 100\% at PIRs of 1\%, 50\%, and 70\%. Furthermore, our analysis of Table \ref{table1} reveals two significant findings. \textbf{Firstly}, when utilizing all labeled data, the instance classification performance is relatively poor at extremely low PIRs (1\%) or high PIRs (50\% and 70\%). This suggests that the task becomes exceptionally challenging at a PIR of 1\% and that, at PIRs of 50\% and 70\%, while the bag classification AUCs reach 1, the instance classification performance notably deteriorates. This observation indicates that accurately identifying all positive instances is not necessary for correctly classifying bags. As the PIR increases, positive bags contain more positive instances, simplifying the bag classification task. The network only needs to identify the simplest positive instance to complete bag classification (assigning the highest attention score to the simplest positive instance in the bag), leading to a decreased motivation to accurately classify all positive instances and exhibiting a "lazy" behavior. \textbf{Secondly}, almost all semi-supervised methods with a 50\% labeled data ratio demonstrate instance classification performance that is comparable to or surpasses the performance of using the entire dataset at PIRs of 1\%, 50\%, and 70\%. This can be attributed to the utilization of a substantial amount of unlabeled data in the WSWC task for consistency constraints, which introduces additional difficulty to bag classification and mitigates the "lazy" behavior of the attention module to some extent.

\subsection{Results on the Camelyon 16 Dataset}

The results obtained on the Camelyon 16 Dataset are presented in Table \ref{table2}. Remarkably, CroCo consistently demonstrates superior performance in both bag-level and instance-level classification, outperforming all compared methods by a significant margin. Notably, even with a labeled data ratio of only 50\%, CroCo achieves bag classification performance comparable to that of the Full-data scenario (100\% labeled data ratio). Furthermore, with a mere 10\% labeled data ratio, CroCo surpasses the instance classification performance of the Full-data scenario.

\subsection{Results on the TCGA-Lung Cancer Dataset and the Clinical Cervical Dataset}

The results the TCGA-Lung Cancer Dataset are shown in Table \ref{table3}. CroCo achieves the best bag-level classification performance under all labeled data ratios. 

The results on the Clinical Cervical Dataset are shown in Table \ref{table4}. As introduced in the Supplementary Materials, this is a more difficult task than the previous two real-world datasets. CroCo still achieved the best bag classification performance and significantly outperformed all compared methods. Using only 50\% of the labeled data, CroCo's performance was already very close to Full-data. In clinical practice, acquiring prognosis labels typically requires long-term and meticulous follow-up, which often poses challenges. In this regard, CroCo proves to be valuable by reducing the labeling requirements by half, thereby offering assistance in overcoming such obstacles.

\begin{table*}[t!]
  \begin{minipage}{\textwidth}
    \centering
    \caption{The performance of bag-level classification on the TCGA-Lung Cancer Dataset. The bag-based classifier branch used ABMIL \cite{8} and DSMIL \cite{13} as backbones respectively. Full-data represents training the corresponding bag-based classifier using 100\% labeled data ratio.}
    \label{table3}

    \subcaptionbox{Performance on ABMIL backbone.}{%
      \begin{adjustbox}{max width=0.7\textwidth}
        \begin{tabular}{c|cccccc}
          \hline
          Backbone:ABMIL & \multicolumn{6}{c}{Performance of bag-level classification} \\ \hline
          labeled data ratio & 10\% & 20\% & 30\% & 40\% & \multicolumn{1}{c|}{50\%} & Full-data \\ \hline
          Baseline & 0.8601 & 0.8708 & 0.9017 & 0.9168 & \multicolumn{1}{l|}{0.9303} & \\
          MT (17'NeurIPS) & 0.8793 & 0.8920 & 0.9171 & 0.9242 & \multicolumn{1}{l|}{0.9360} & \\
          Fixmatch (20'NeurIPS) & 0.8876 & 0.8969 & 0.9254 & 0.9286 & \multicolumn{1}{l|}{0.9376} & \\
          Freematch (22'ICLR) & 0.8973 & 0.9064 & 0.9251 & 0.9328 & \multicolumn{1}{l|}{0.9415} & \\
          \textbf{CroCo (Ours)} & {\color[HTML]{FF0000}\textbf{0.9063}} & {\color[HTML]{FF0000}\textbf{0.9126}} & {\color[HTML]{FF0000}\textbf{0.9318}} & {\color[HTML]{FF0000}\textbf{0.9387}} & \multicolumn{1}{l|}{{\color[HTML]{FF0000}\textbf{0.9462}}} & \multirow{-5}{*}{{\color[HTML]{656565} \textbf{0.9488}}} \\ \hline
        \end{tabular}%
      \end{adjustbox}
    }

    \subcaptionbox{Performance on DSMIL backbone.}{%
      \begin{adjustbox}{max width=0.7\textwidth}
        \begin{tabular}{c|cccccc}
          \hline
          Backbone:DSMIL & \multicolumn{6}{c}{Performance of bag-level classification} \\ \hline
          labeled data ratio & 10\% & 20\% & 30\% & 40\% & \multicolumn{1}{c|}{50\%} & Full-data \\ \hline
          Baseline & 0.8625 & 0.8716 & 0.9094 & 0.9271 & \multicolumn{1}{c|}{0.9362} & \\
          MT (17'NeurIPS) & 0.8744 & 0.8895 & 0.9277 & 0.9442 & \multicolumn{1}{c|}{0.9573} & \\
          Fixmatch (20'NeurIPS) & 0.8742 & 0.8993 & 0.9365 & 0.9467 & \multicolumn{1}{c|}{0.9599} & \\
          Freematch (22'ICLR) & 0.8831 & 0.9056 & 0.9374 & 0.9470 & \multicolumn{1}{c|}{0.9602} & \\
          \textbf{CroCo (Ours)} & {\color[HTML]{FF0000}\textbf{0.9121}} & {\color[HTML]{FF0000}\textbf{0.9257}} & {\color[HTML]{FF0000}\textbf{0.9436}} & {\color[HTML]{FF0000}\textbf{0.9584}} & \multicolumn{1}{c|}{{\color[HTML]{FF0000}\textbf{0.9612}}} & \multirow{-5}{*}{{\color[HTML]{656565} \textbf{0.9633}}} \\ \hline
        \end{tabular}%
      \end{adjustbox}
    }
  \end{minipage}
\end{table*}

\begin{table*}[t!]
  \centering
  \caption{The performance of bag-level classification on the Clinical Cervical Dataset. The bag-based classifier branch used ABMIL \cite{8} and DSMIL \cite{13} as backbones respectively. Full-data represents training the corresponding bag-based classifier using 100\% labeled data ratio.}
  \label{table4}

  \subfloat[Performance on ABMIL backbone.]{%
    \begin{adjustbox}{max width=0.7\textwidth}
    \begin{tabular}{c|cccccc}
   \hline
    Backbone:ABMIL          & \multicolumn{6}{c}{Performance of   bag-level classification}                                                                                                                                                                                                                              \\ \hline
    labeled data ratio             & 10\%                                   & 20\%                                    & 30\%                                    & 40\%                                    & \multicolumn{1}{c|}{50\%}                                    & Full-data                                                \\ \hline
Baseline                & 0.4500                                 & 0.5004                                 & 0.5422                                 & 0.6244                                 & \multicolumn{1}{c|}{0.6533   }                              & {\color[HTML]{808080} }                                  \\
MT   (17'NeurIPS)       & 0.5669                                 & 0.6454                                 & 0.6576                                 & 0.6872                                 & \multicolumn{1}{c|}{0.6822  }                               & {\color[HTML]{808080} }                                  \\
Fixmatch   (20'NeurIPS) & 0.6230                                 & 0.6559                                 & 0.6916                                 & 0.7044                                 & \multicolumn{1}{c|}{0.7000 }                                & {\color[HTML]{808080} }                                  \\
Freematch   (22'ICLR)   & 0.6446                                 & 0.6791                                 & 0.6999                                 & 0.7156                                 & \multicolumn{1}{c|}{0.7222}                                 & {\color[HTML]{808080} }                                  \\
\textbf{CroCo   (Ours)}  & {\color[HTML]{FF0000} \textbf{0.6754}} & {\color[HTML]{FF0000} \textbf{0.6889}} & {\color[HTML]{FF0000} \textbf{0.7122}} & {\color[HTML]{FF0000} \textbf{0.7267}} & \multicolumn{1}{c|}{\color[HTML]{FF0000} \textbf{0.7389}} & \multirow{-5}{*}{{\color[HTML]{808080} \textbf{0.7439}}} \\ \hline
    \end{tabular}%
    \end{adjustbox}
  }

  \subfloat[Performance on DSMIL backbone.]{%
    \begin{adjustbox}{max width=0.7\textwidth}
    \begin{tabular}{c|cccccc}
    \hline
    Backbone:DSMIL          & \multicolumn{6}{c}{Performance of   bag-level classification}                                                                                                                                                                                                                              \\ \hline
    labeled data ratio             & 10\%                                   & 20\%                                    & 30\%                                    & 40\%                                    & \multicolumn{1}{c|}{50\%}                                    & Full-data                                                \\ \hline
Baseline                & 0.4222                                 & 0.4811                                 & 0.5543                                 & 0.6233                                 &\multicolumn{1}{c|}{ 0.6825}                                 & {\color[HTML]{808080} }                                  \\
MT   (17'NeurIPS)       & 0.5324                                 & 0.5896                                 & 0.6323                                 & 0.6752                                 & \multicolumn{1}{c|}{0.7021}                                 & {\color[HTML]{808080} }                                  \\
Fixmatch   (20'NeurIPS) & 0.6522                                 & 0.6722                                 & 0.6967                                 & 0.7200                                 & \multicolumn{1}{c|}{0.7444}                                 & {\color[HTML]{808080} }                                  \\
Freematch   (22'ICLR)   & 0.6644                                 & 0.6836                                 & 0.7144                                 & 0.7372                                 & \multicolumn{1}{c|}{0.7522}                                 & {\color[HTML]{808080} }                                  \\
\textbf{CroCo   (Ours)}  & {\color[HTML]{FF0000} \textbf{0.6756}} & {\color[HTML]{FF0000} \textbf{0.6911}} & {\color[HTML]{FF0000} \textbf{0.7389}} & {\color[HTML]{FF0000} \textbf{0.7589}} & \multicolumn{1}{c|}{\color[HTML]{FF0000} \textbf{0.7718}} & \multirow{-5}{*}{{\color[HTML]{808080} \textbf{0.7932}}} \\ \hline
    \end{tabular}%
    \end{adjustbox}
  }

\end{table*}

\subsection{Ablation Study and Key Advantages Analysis}
\textbf{The key advantages of CroCo} are: 1) the use of a two-level cross-consistency strategy, which fully utilizes both instance-level and bag-level information. 2) the use of a heterogeneous dual-branch design, which fully integrates the advantages of bag-based classifiers and instance-based classifiers, and allows them to mutually promote each other. We conducted ablation experiments on the Camelyon 16 dataset at 10\%, 30\%, and 50\% labeled data ratios targeting these two points.

\textbf{Two-level Consistency} and \textbf{Bi-directional Cross-supervision}. Table \ref{table5} (a) shows the ablation results on the two-level consistency (first three rows) and the bi-directional cross-supervision (last three rows) designs. \textbf{From the first three rows}, we can see that using two-level cross consistency is significantly better than using either one of them, especially when the labeled data ratio is low. \textbf{From the last three rows}, we can see that CroCo's bi-directional cross-supervision achieves the best bag-level and instance-level performance, superior to either single-direction consistency supervision. 

\textbf{Heterogeneous Dual Branches, True Negative Instances and Shared Encoder}. Table \ref{table5} (b) shows the ablation results of CroCo's heterogeneous dual branches (first three rows) and other framework design choices (last three rows). \textbf{The data in the first three rows} shows that using homogeneous dual branches of bag-based classifiers or instance-based classifiers does not perform as well as using heterogeneous dual branches. Compared with using homogeneous dual branches of instance-based classifiers, using homogeneous dual branches of bag-based classifiers can achieve better performance in bag-level classification. Conversely, using homogeneous instance-based classifiers can achieve better performance in instance-level classification, which also reflects the different advantages of the two kinds of classifiers in bag-level and instance-level classification. CroCo uses a heterogeneous dual-branch architecture design, which achieves the best bag-level and instance-level classification performance. On one hand, the heterogeneous dual branches avoids the degradation of the homogeneous dual-branch structure, and on the other hand, it fully integrates the advantages of bag-based classifiers and instance-based classifiers and enables them to guide and promote each other. \textbf{The data in the last three rows} shows that CroCo fully utilized true negative instances from negative bags when constructing heterogeneous dual branches, and shared the instance-level encoder parameters of the two branches. The former can effectively promote instance-level training, while the latter can facilitate the transfer and promotion of knowledge between the two branches.

\begin{table*}[t!]
  \centering
  \caption{Results of the ablation experiments. "bag" and "instance" denote consistency supervision at the bag and the instance levels, respectively. "bag cls.→ins cls." signifies the utilization of the bag-based classifier solely for supervising the instance-based classifier at both the instance and bag levels. Conversely, "ins cls.→bag cls." implies the utilization of the instance-based classifier solely for supervising the bag-based classifier at both the instance and bag levels. "bag cls." and "ins cls." refer to the bag-based and instance-based classifiers, respectively. The percentage indicates the labeled data ratio.}

  \subfloat[Ablation experiments on the two-level consistency (first three rows) and bi-directional cross-supervision (last three rows) designs.]{%
    \begin{adjustbox}{max width=0.7\textwidth}
      \begin{tabular}{c|ccc|ccc}
      \hline
      \multirow{2}{*}{Unlabeled supervision}    & \multicolumn{3}{c|}{Bag-level performance} & \multicolumn{3}{c}{Instance-level performance} \\ \cline{2-7} 
                                                & 10\%         & 30\%         & 50\%         & 10\%           & 30\%           & 50\%           \\ \hline
      bag only                                  & 0.6836       & 0.7465       & 0.7898       & 0.8157         & 0.8334         & 0.8742         \\
      ins only                                  & 0.6790       & 0.7323       & 0.7742       & 0.7907         & 0.8435         & 0.8905         \\
      \textbf{bag + ins (ours)}                 & \color[HTML]{FF0000}\textbf{0.7688}       & \color[HTML]{FF0000}\textbf{0.8025}       & \color[HTML]{FF0000}\textbf{0.8352}       & \color[HTML]{FF0000}\textbf{0.8739}       & \color[HTML]{FF0000}\textbf{0.9059}       & \color[HTML]{FF0000}\textbf{0.9127}     \\
      \hline
      bag + ins (bag cls.→ins cls.)             & 0.7179       & 0.7655       & 0.8141       & 0.8275         & 0.8428         & 0.8933         \\
      bag + ins (ins cls.→bag cls.)             & 0.7018       & 0.7528       & 0.8059       & 0.8432         & 0.8665         & 0.9106         \\
      \textbf{bag + ins (ours)}                 & \color[HTML]{FF0000}\textbf{0.7688}       & \color[HTML]{FF0000}\textbf{0.8025}       & \color[HTML]{FF0000}\textbf{0.8352}       & \color[HTML]{FF0000}\textbf{0.8739}       & \color[HTML]{FF0000}\textbf{0.9059}       & \color[HTML]{FF0000}\textbf{0.9127}         \\ \hline
      \end{tabular}
    \end{adjustbox}
  }
  \hfill
  \subfloat[Ablation experiments on the heterogeneous architecture design (first three rows) and other network details (last three rows).]{%
    \begin{adjustbox}{max width=0.7\textwidth}
      \begin{tabular}{c|ccc|ccc}
      \hline
      \multirow{2}{*}{Branch}                 & \multicolumn{3}{c|}{Bag-level performance} & \multicolumn{3}{c}{Instance-level performance} \\ \cline{2-7} 
                                              & 10\%         & 30\%         & 50\%         & 10\%           & 30\%           & 50\%           \\ \hline
      bag cls. + bag cls.                     & 0.7527       & 0.7835       & 0.7995       & 0.7753         & 0.8046         & 0.8696         \\
      ins cls. + ins cls.                     & 0.6415       & 0.7485       & 0.7738       & 0.8265         & 0.8533         & 0.8913         \\
      \textbf{bag cls. + ins cls. (ours)}     & \color[HTML]{FF0000}\textbf{0.7688}       & \color[HTML]{FF0000}\textbf{0.8025}       & \color[HTML]{FF0000}\textbf{0.8352}       & \color[HTML]{FF0000}\textbf{0.8739}       & \color[HTML]{FF0000}\textbf{0.9059}       & \color[HTML]{FF0000}\textbf{0.9127}          \\
      \hline
      bag cls. + ins cls. (w/o TrueNeg)       & 0.7418       & 0.7864       & 0.8209       & 0.8570         & 0.8722         & 0.9084         \\
      bag cls. + ins cls. (w/o ShareEncoder)  & 0.7210       & 0.7442       & 0.7915       & 0.8457         & 0.8655         & 0.9008         \\
      \textbf{bag cls. + inst cls. (ours)}    & \color[HTML]{FF0000}\textbf{0.7688}       & \color[HTML]{FF0000}\textbf{0.8025}       & \color[HTML]{FF0000}\textbf{0.8352}       & \color[HTML]{FF0000}\textbf{0.8739}       & \color[HTML]{FF0000}\textbf{0.9059}       & \color[HTML]{FF0000}\textbf{0.9127}         \\
      \hline
      \end{tabular}
    \end{adjustbox}
  }

  \label{table5}%
\end{table*}

\begin{figure*}[t!]
  \centering
  \includegraphics[width=\textwidth]{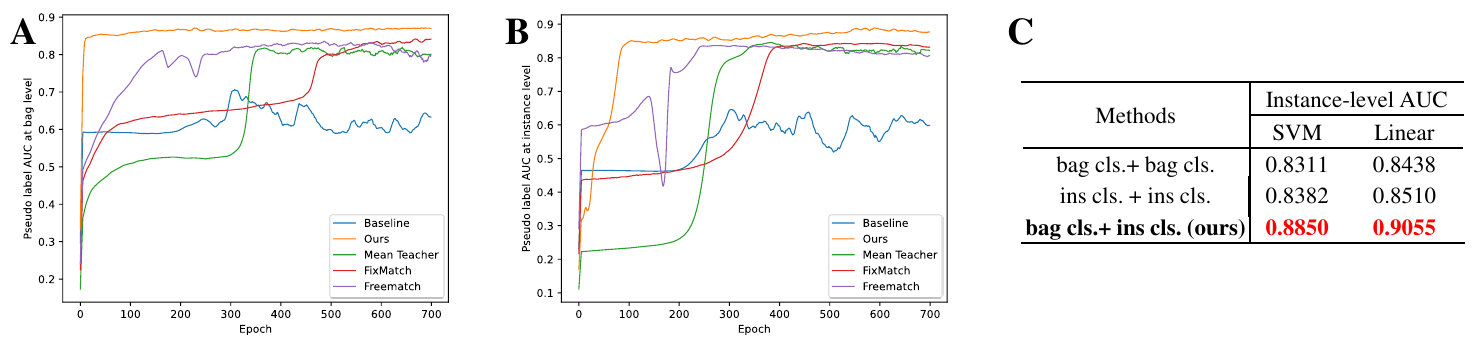}
  \caption{The AUC curves of CroCo's bag-level pseudo labels (A) and instance-level pseudo labels (B) during training. (C) SVM and linear evaluation of extracted features on the Camelyon16 Dataset.}
  \label{figure3}
\end{figure*}

\textbf{Feature Extraction of the Heterogeneous Dual Branches}. We further illustrate the efficiency of the heterogeneous dual-branch structure adopted by CroCo from the perspective of features. First, we use the instance-level feature extractor trained using two homogeneous bag-based classifiers, two homogeneous instance-based classifiers, and our heterogeneous dual-branch model to extract features for all instances from the training set and the test set. Note that all methods use the ResNet-18 as the instance encoder for feature extraction. Then, we use the true labels of each instance to train a simple SVM classifier and a linear classifier on the training set. Finally, we evaluate these two classifiers on the test set. The results are shown in Figure \ref{figure3} C. It can be seen that the instance features extracted using our heterogeneous dual-branch architecture achieve the highest performance, indicating that our method obtains the best feature space.

\textbf{AUC of Bag-level and Instance-level Pseudo Labels}. We conducted a detailed analysis of the AUC curves for the bag-level pseudo labels and instance-level pseudo labels during the training of CroCo. This analysis showcases CroCo's ability to effectively leverage unlabeled data from both levels. Figure \ref{figure3} A and B visually demonstrate the superior performance of CroCo compared to the comparative methods in terms of convergence speed, final indicators (AUC), and the stability of pseudo labels for both bag-level and instance-level tasks. The higher quality of the pseudo labels, along with the faster and more stable training process, indicate that CroCo excels in harnessing the potential of unlabeled data from both levels. Consequently, CroCo enables the learning of a more accurate and stable feature space.

\section{Conclusion}
In this paper, we introduce a new Weakly Semi-supervised WSI Classification (WSWC) problem setting that is better in line with clinical reality: a task that involves both labeled (probably fewer) and unlabeled bags in WSI classification. We design a concise and efficient cross consistency supervision framework for both bag level and instance level, called CroCo, to effectively solve this problem. Results from extensive experiments on four datasets show that, CroCo achieves better bag and instance classification performance compared to other comparative methods.
However, our paper is limited by the absence of an additional filtering strategy to pseudo-labeling, potentially influenced by inaccurate pseudo-labels. Moreover, CroCo effectively minimizes the requirement for data annotation in a clinical context, assisting in overcoming challenges like data acquisition.

\bibliographystyle{fcs}
\bibliography{fcs}

\end{document}